\documentclass[lettersize,journal]{IEEEtran}
\usepackage{amsmath,amsfonts}
\usepackage{algorithm}
\usepackage{array}
\usepackage[caption=false,font=normalsize,labelfont=sf,textfont=sf]{subfig}
\usepackage{textcomp}
\usepackage{stfloats}
\usepackage{url}
\usepackage{verbatim}
\usepackage{graphicx}
\usepackage{cite}

\usepackage{hyperref}
\usepackage{booktabs,multirow}
\usepackage{makecell}
\usepackage{bbm}
\usepackage{enumitem}
\usepackage{algpseudocode}
\usepackage{tabularx}

\usepackage[table]{xcolor} 
\definecolor{WinGreen}{RGB}{100,210,100}
\definecolor{RunGreen}{RGB}{200,235,200}
\newcommand{\win}[1]{\cellcolor{WinGreen}\textbf{#1}}
\newcommand{\ru}[1]{\cellcolor{RunGreen}\underline{#1}}

\definecolor{RevisionBlue}{RGB}{0,0,0}
\DeclareRobustCommand{\revisedtext}[1]{{\color{RevisionBlue}#1}}

\begin{document}

\title{Label-Efficient Document-Grounded Factuality Detection via Cross-Paraphrastic Invariance Learning}

\author{Chao Chen, Shanshan Lin, Dongsheng Hong, Sibo Ju, Wenjie Yang, Sihong Xie,~\IEEEmembership{Senior Member,~IEEE}, and Xiangwen Liao%
\thanks{Chao Chen is with Harbin Institute of Technology (Shenzhen).}%
\thanks{Shanshan Lin, Dongsheng Hong, Sibo Ju, Wenjie Yang, and Xiangwen Liao are with Fuzhou University.}%
\thanks{Sihong Xie is with Hong Kong University of Science and Technology (Guangzhou).}%
\thanks{Xiangwen Liao is the corresponding author.}
\thanks{\textcopyright{} 2026 IEEE.
Personal use of this material is permitted.  Permission from IEEE must be obtained for all other uses, in any current or future media, including reprinting/republishing this material for advertising or promotional purposes, creating new collective works, for resale or redistribution to servers or lists, or reuse of any copyrighted component of this work in other works.}
}



\maketitle

\begin{abstract}
Large language models (LLMs) can generate fluent claims that are unsupported by a given source document, motivating accurate and efficient document-grounded factuality detection.
Existing LLM-based evaluators often require repeated inference and incur high computational cost, 
whereas compact factuality detectors are efficient at inference time but commonly rely on large labeled or synthetically generated corpora.
We propose \textbf{CPIL} (\emph{Cross-Paraphrastic Invariance Learning}), a two-stage Siamese framework that converts a small labeled set into relational supervision without acquiring additional factuality labels. 
CPIL constructs positive pairs from distinct label-preserving paraphrastic views of the same document-claim instance and negative pairs from opposite-label examples associated with the same source document. 
It further introduces quantile-aware semi-hard negative mining, which derives a boundary from the positive-pair similarity distribution and balances informative negatives with broader coverage. 
In the two-stage training procedure, 
Stage~1 learns an encoder from these task-aware pairs,
and Stage~2 jointly fine-tunes the encoder and a linear classification head for binary factuality prediction.
At inference time, CPIL requires no paraphrase generation, pair construction, or Siamese comparison and uses only one encoder-classifier forward pass. 
On the 11-task \textit{LLM-AggreFact} benchmark, CPIL-2\% and CPIL-3\% achieve macro-F1 scores of 81.04 and 81.60 while using only 2-3\% of the FactCG-released labeled pool.
Both settings outperform all compared compact detectors and two vanilla LLM references.
Ablations and sensitivity analyses support the contributions of task-aware pair construction, dynamic negative mining, and multi-pivot paraphrasing.
\end{abstract}


\section{Introduction}
\label{sec:intro}
Large language models (LLMs) can produce fluent and coherent text, but fluency does not guarantee faithfulness to the available evidence.
A common failure mode is \emph{hallucination}, where generated claims are inconsistent with or unsupported by a provided source document~\cite{rahman2025hallucination_review,maynez2020faithfulness}. 
Detecting such unsupported claims is therefore important for building trustworthy and reliable generation systems~\cite{krishna2025safety}.

\begin{figure}
    \centering
    \includegraphics[width=.95\linewidth]{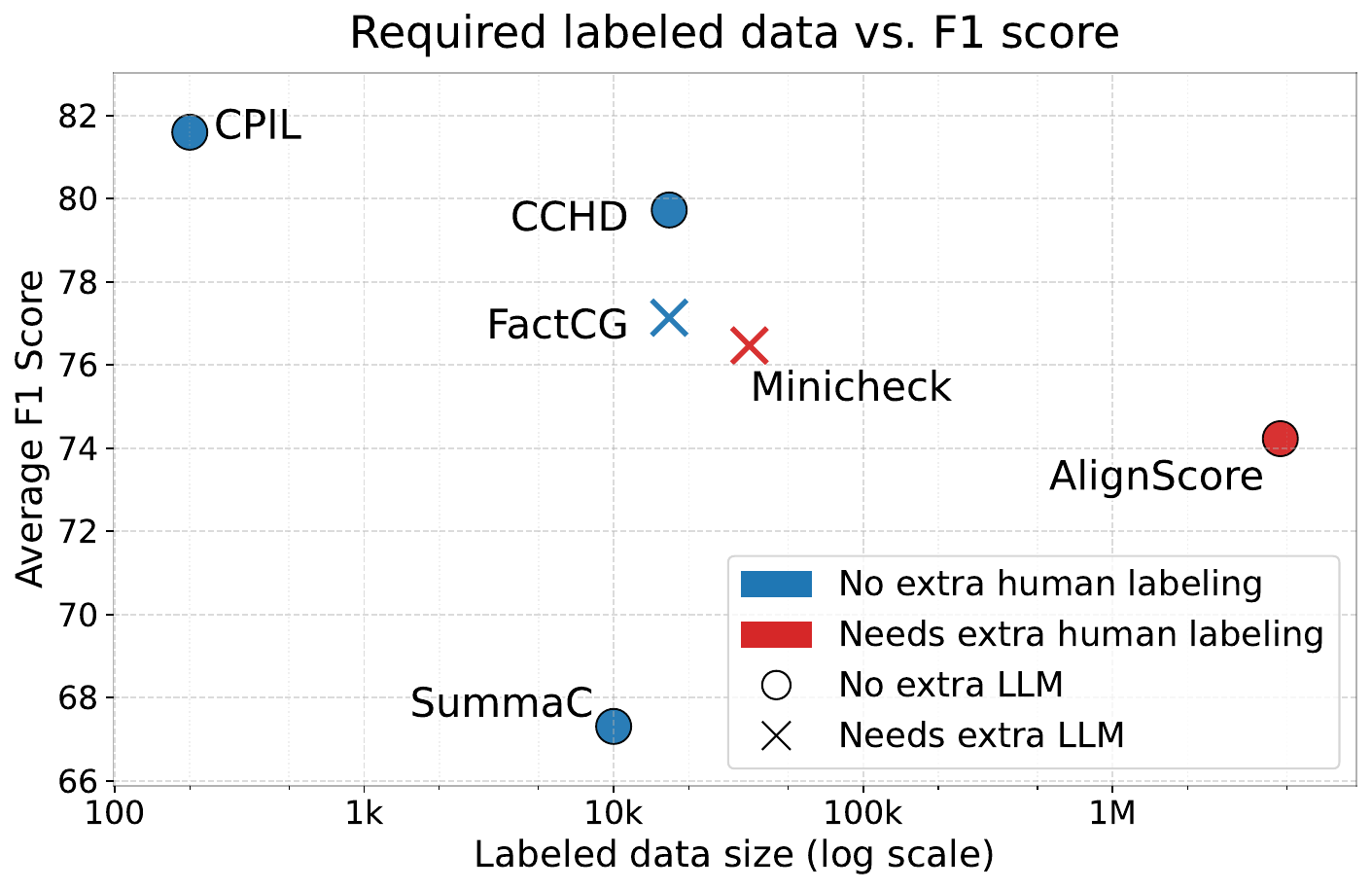}
    \caption{
    Required labeled data size \textit{vs.} model performance in macro-F1.
    Colors and markers indicate whether additional human or LLM-based labeling is required, respectively.
    }
    \label{fig:labeled_data_f1}
\end{figure}

Existing factuality detectors often trade off inference cost against supervision cost~\cite{rahman2025hallucination_review}.
LLM-as-evaluator methods, including SelfCheckGPT~\cite{manakul2023selfcheckgpt}, SelfCheckAgent~\cite{muhammed2025selfcheckagent}, and G-Eval~\cite{liu2023geval}, ask powerful LLMs to judge or verify generated content.
These methods are flexible, but evaluation may involve repeated prompting, long-context reasoning, or multi-step verification, leading to substantial inference cost~\cite{tang2024minicheck}.
Compact factuality detectors are more efficient at deployment because they directly predict whether a claim is supported by a document.
Representative methods include QAFactEval~\cite{fabbri2021qafacteval}, which uses question answering; AlignScore~\cite{zha2023alignscore}, which learns alignment scores from multiple supervision sources; MiniCheck~\cite{tang2024minicheck}, which trains on LLM-labeled document-claim pairs; and FactCG~\cite{lei2025factcg}, which augments training data with graph-generated multi-hop claims.
As summarized in Fig.~\ref{fig:labeled_data_f1}, however, many competitive compact detectors still rely on substantial human-labeled or synthetically generated data~\cite{zha2023alignscore,tang2024minicheck,lei2025factcg}.
The challenge is therefore not only to build an accurate detector, 
but also to reduce its factuality-label requirement while preserving efficient inference.

We address this challenge with \textbf{CPIL} (\emph{Cross-Paraphrastic Invariance Learning}), a two-stage Siamese learning framework that converts a small labeled subset into task-aware relational supervision.
The key motivation is that standard instance-wise classification extracts one supervision target from each labeled document-claim example, whereas the same labels can induce structured constraints over multiple views and comparisons.
CPIL converts limited instance-level supervision into relational constraints through three mechanisms.

First, for \emph{positive supervision}, CPIL generates back-translated views of each document-claim instance and explicitly aligns different views of the same example.
Unlike append-only augmentation, this cross-view constraint encourages invariance to assumed label-preserving lexical and syntactic variation.
Second, for \emph{negative supervision}, CPIL contrasts opposite-label examples associated with the same source document.
These same-document negatives reduce direct topical and domain shortcuts and encourage the encoder to model finer differences in the document-claim support relation.
Third, CPIL introduces \emph{quantile-aware semi-hard negative mining} to adapt negative difficulty as the representation space evolves. 
It derives a boundary from the current positive-pair similarity distribution, filters overly positive-like candidates, and combines informative semi-hard negatives with broader coverage from the remaining safe pool.

The detector is optimized in two stages.
\textit{Stage~1} pretrains a weight-tied encoder with a contrastive objective over cross-paraphrase positives and dynamically selected same-document negatives.
\textit{Stage~2} attaches a linear classification head and jointly fine-tunes the encoder and classifier on the same low-budget labeled subset. 
Paraphrase generation, pair construction, and dynamic mining are restricted to training.
At inference time, CPIL evaluates each document-claim pair with a single encoder-classifier forward pass.

On the 11-task \textit{LLM-AggreFact} benchmark \cite{tang2024minicheck}, CPIL achieves the highest macro-F1 among the compared compact factuality detectors,
and also exceeds three vanilla LLM references.
CPIL-2\% and CPIL-3\% obtain macro-F1 scores of 81.04 and 81.60, respectively, while using only 2--3\% of the FactCG-released labeled pool.
The results are supported by ablation studies of the pairwise stage, positive and negative pair definitions, and dynamic mining, as well as sensitivity analyses of the label budget, semi-hard negative proportion, pivot-language selection.

The main contributions are summarized as follows:
\begin{itemize}
    \item We introduce a factuality-specific relational supervision strategy that constructs \textit{cross-paraphrase positives} and \textit{same-document, opposite-label negatives}. 
    \item We propose a \textit{quantile-aware semi-hard negative mining} strategy that derives a positive-informed safety boundary and balances high-similarity negative selection with broader coverage. 
    \item We instantiate these mechanisms into CPIL, a \textit{two-stage} Siamese framework that requires neither additional factuality labels nor pairwise
    inference. 
    \item CPIL demonstrates strong low-budget macro-F1 on the 11-task LLM-AggreFact benchmark.
\end{itemize}

\begin{table*}[t]
    \centering
    \caption{Main extensions over the conference paper~\cite{lin2026cpil}.}
    \label{tab:conference_journal_difference}
    \renewcommand{\arraystretch}{1.12}
    \setlength{\tabcolsep}{4pt}
    \begin{tabularx}{\textwidth}{
        >{\raggedright\arraybackslash}p{0.10\textwidth}
        >{\raggedright\arraybackslash}p{0.38\textwidth}
        >{\raggedright\arraybackslash}X}
        \toprule
        \textbf{Aspect}
        & \textbf{Conference paper}
        & \textbf{Journal manuscript} \\
        \midrule

        Method
        &
        Two-stage Siamese learning with \textit{static} cross-paraphrase positives and same-document negatives.
        &
        Adds \textit{quantile-aware semi-hard negative mining} with a positive-derived boundary and an exploitation-coverage strategy.
        \\

        Formulation
        &
        \textit{Concise} formulation of pair construction and two-stage         training.
        &
        \textit{Expanded} formulation with separate offline and dynamic        training algorithms and complexity analysis.
        \\

        Evaluation
        &
        Conference-scale baseline comparison and ablations.
        &
        Stronger baselines and broader ablations of augmentation, pair construction, and dynamic mining.
        \\

        Analysis
        &
        \textit{Limited} low-budget study.
        &
        \textit{Additional} studies of label budgets, mining ratios, pivot languages, and multi-seed analyses.
        \\

        \bottomrule
    \end{tabularx}
\end{table*}

\textbf{Journal-version extension.}
This article substantially extends the conference paper \cite{lin2026cpil} through a new dynamic negative-mining method, expanded method specification, and broader controlled evaluation. Table~\ref{tab:conference_journal_difference} summarizes the main differences between the two versions.


\section{Related Work}
\label{sec:related_work}

\subsection{Document-Grounded Factuality Evaluation}

\paragraph{Compact factuality detectors}
Document-grounded factuality detection evaluates whether a claim is supported by a given evidence document.
Prior factual consistency methods for summarization often formulate this problem as natural language inference (NLI) or question answering (QA).
FactCC trains a weakly supervised factual consistency classifier using synthetically perturbed source-summary pairs~\cite{kryscinski2019factcc};
SummaC aggregates sentence-level NLI signals for summary consistency~\cite{laban2022summac};
and QuestEval and QAFactEval assess consistency through question generation and question answering~\cite{scialom2021questeval,fabbri2021qafacteval}.
More recent compact detectors target broader document-grounded generation settings.
AlignScore learns a unified alignment function from multiple supervision sources~\cite{zha2023alignscore};
MiniCheck trains efficient fact-checkers from synthetically generated document-claim examples~\cite{tang2024minicheck};
FactCG strengthens factuality checkers with graph-based multi-hop synthetic claims~\cite{lei2025factcg};
and CCHD studies constrained paraphrase consistency for factuality detection~\cite{lin2026cchd}. 
Besides, complementary generation-side work has incorporated structured factual knowledge directly into generation~\cite{mao2022fact}. 
However, many competitive compact detectors rely on substantial annotated, synthetic, or model-generated supervision. 
Among these methods, CCHD is the closest to CPIL because both use paraphrastic consistency. 
The two methods, however, impose this consistency at different stages: 
CCHD imposes consistency constraints during classification, whereas CPIL first converts paraphrastic views and same-document label contrasts into representation-level pairwise supervision.  

\paragraph{LLM-based evaluators}
LLM-based evaluators constitute another major line of factuality assessment.
SelfCheckGPT detects hallucinations from inconsistencies among sampled model responses~\cite{manakul2023selfcheckgpt};
G-Eval uses instruction-following LLMs for generation evaluation~\cite{liu2023geval};
FacTool augments factuality assessment with external tools across multiple tasks and domains~\cite{chern2023factool};
and SAFE evaluates long-form factuality by decomposing responses into individual facts and verifying them through search-augmented reasoning~\cite{wei2024long}.
\revisedtext{AFEV iteratively decomposes complex claims into atomic facts and combines evidence retrieval with adaptive reasoning. This inference-time decomposition is complementary to CPIL, which learns from training-time relations and predicts support for a supplied document-claim pair~\cite{zheng2026afev}.}
Beyond post-hoc evaluation, recent work \cite{lin2025variational} mitigates hallucination during generation with latent prompt learning.
Such evaluators can support flexible natural-language reasoning, but their inference procedures may depend on repeated model calls, external retrieval or tools, model access, and prompt design.
In contrast, CPIL preserves a single compact encoder-classifier pass at inference time.

\paragraph{Evaluation benchmarks}
Evaluation resources have also expanded to broad document-grounded settings.
TRUE consolidates multiple factual consistency datasets across summarization and dialogue settings~\cite{honovich2022true};
LLM-AggreFact aggregates diverse datasets for document-grounded hallucination evaluation~\cite{tang2024minicheck};
RAGTruth provides manually annotated hallucination labels for retrieval-augmented generation at both response and word levels~\cite{niu2024ragtruth};
and FActScore evaluates long-form generations by decomposing them into atomic facts and checking factual precision against external evidence~\cite{min2023factscore}.
Together, these benchmarks emphasize fine-grained grounding, evidence sensitivity, and cross-domain generalization.

\subsection{Contrastive, Siamese, and Label-Efficient Learning}
Contrastive and Siamese methods are attractive under limited supervision because existing examples can induce multiple relational learning signals. 
SimCLR~\cite{chen2020simclr} defines positives through data augmentation, supervised contrastive learning~\cite{khosla2020contrastive_learning,gunel2020supervised} organizes relations using class labels, and Siamese networks~\cite{he2018siamese,chen2021snx} employ weight-tied encoders to compare paired inputs in a shared representation space. 
MultiSCL~\cite{li2023multi} further combines sentence-level and pair-level supervised contrastive objectives with augmented views to learn discriminative representations from limited labels. 
However, generic pair definitions can be poorly aligned with document-grounded factuality: cross-document negatives may be separable through topic or domain mismatch, while unrelated same-label examples need not rely on comparable evidence and can form weak positives.
CPIL therefore ties pair construction to the factuality task itself by aligning views of the same document-claim instance and contrasting opposite-label claims under the same document.

\subsection{Paraphrase Augmentation, Invariance, and Hard Negatives}
Paraphrase augmentation has been widely used to improve robustness and data efficiency through lexical perturbation \cite{wei2019eda}, back-translation \cite{sennrich2016improving,edunov2018understanding}, and paraphrase generation~\cite{beddiar2021data,agarwal2023perturb,yadav2024pag}. 
Unsupervised paraphrase generation and information-dropping augmentation provide further examples of data-efficient text transformation~\cite{li2024unsupervised,li2024information}. 
For factuality detection, however, paraphrastic transformations must preserve not only broad semantics but also the original support relation, since changes to entities, quantities, temporal expressions, or negation may alter the label.
Moreover, append-style augmentation exposes a classifier to additional surface forms but does not explicitly constrain corresponding views to have similar representations; 
CPIL instead uses cross-view alignment to learn paraphrastic invariance.
Hard negatives can further sharpen contrastive learning~\cite{schroff2015facenet,khosla2020contrastive_learning,gunel2020supervised}, but difficulty must be controlled carefully: uniformly sampled negatives may become easy as training progresses, whereas the most similar opposite-label pairs may be ambiguous, noisy, or overly positive-like.
CPIL therefore uses the positive-pair similarity distribution to filter unsafe candidates and balance informative negatives with broader coverage.

\begin{figure*}
    \centering
    \includegraphics[width=.95\linewidth]{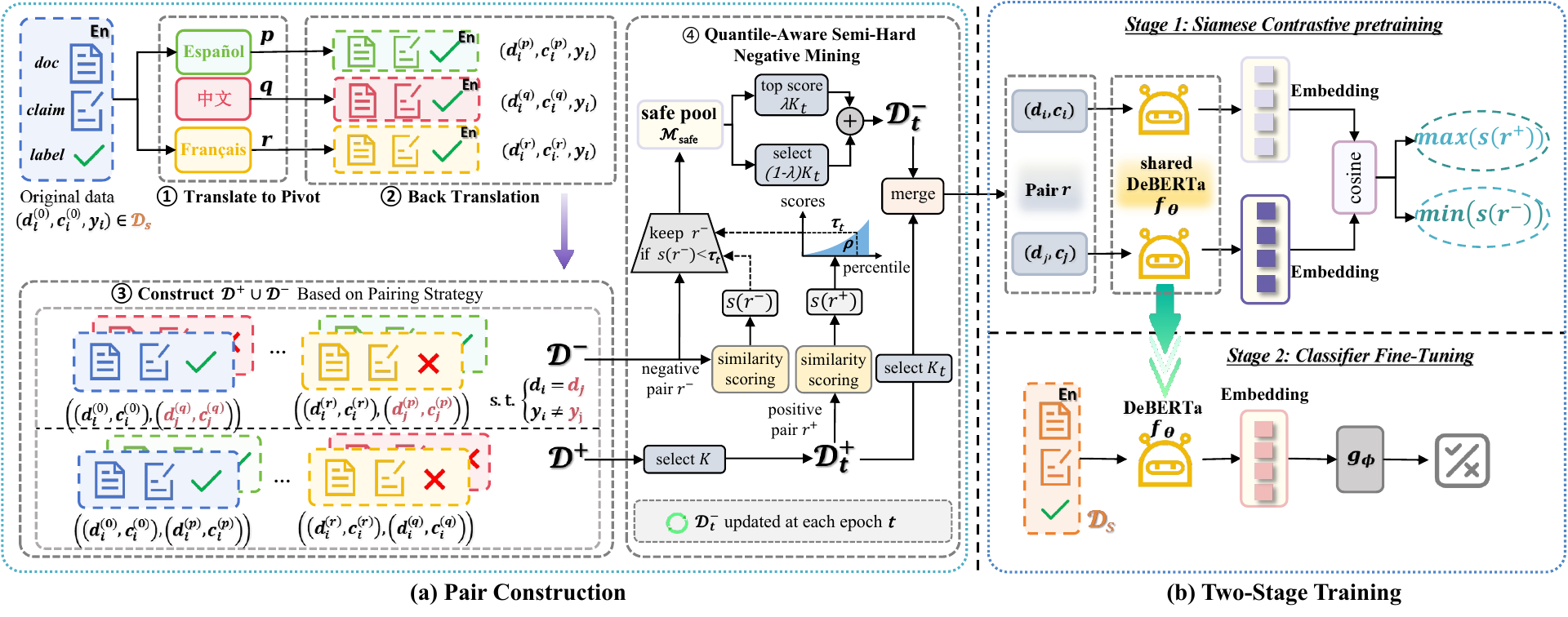}
    \caption{Framework of CPIL.
    \textbf{(a) Pair Construction}: 
    CPIL generates paraphrastic views via back-translation and constructs factuality-aware training pairs.
    Positive pairs are cross-paraphrase views of the same document-claim instance,
    while negative pairs are opposite-label claims grounded in the same source document.
    \textbf{(b) Two-Stage Training}: 
    Stage~1 trains a shared encoder $f_\theta$ with a Siamese pair-similarity ranking objective. 
    Stage~2 attaches a lightweight classifier $g_\phi$ and fine-tunes the detector for binary factuality detection.
    }
    \label{fig:framework}
\end{figure*}

\section{Proposed Method}
\label{sec:method}

This section formalizes document-grounded factuality detection and presents the proposed CPIL, as illustrated in Fig.~\ref{fig:framework}.
Sec.~\ref{sec:prelim} defines the task and notation.
Sec.~\ref{sec:views} introduces the label-preserving paraphrastic views and the factuality-aware positive and negative candidate pairs.
Sec.~\ref{sec:semi_hard_mining} presents the quantile-aware semi-hard negative mining strategy.
Sec.~\ref{sec:training} describes the two-stage optimization procedure, and Sec.~\ref{sec:complexity} discusses its computational complexity.

\subsection{Problem Setup}
\label{sec:prelim}
We formulate document-grounded factuality detection as a binary classification problem~\cite{tang2024minicheck}.
Let \(\mathcal{D}=\{(d_i,c_i,y_i)\}_{i=1}^{N}\) denote the available labeled dataset, where $d_i$ is a source document, $c_i$ is a claim to be verified against $d_i$, and $y_i\in\{0,1\}$ is the factuality label.
We use $y_i=1$ to denote that the claim is \textit{supported} by the document and $y_i=0$ to denote that the claim is \textit{unsupported}.
A factuality detector estimates
\(
h(d,c)=\Pr(y=1\mid d,c).
\)

CPIL focuses on reusing an existing labeled subset \(\mathcal{D}_s\subset\mathcal{D}\) without acquiring additional human factuality labels.
A direct classifier uses each example in $\mathcal{D}_s$ independently and consequently extracts only one instance-level supervision signal from each label.
A generic contrastive strategy could create positives from same-label examples and negatives from different-label examples~\cite{chen2020simclr}.
However, such pair definitions are not well aligned with document-grounded factuality.
Two supported claims from different documents may share the same label but rely on unrelated evidence, while two claims from different documents with opposite labels may be separable through topic, length, or writing style rather than groundedness. 
CPIL thus constructs pairs around the grounding relation, using \emph{cross-paraphrase positives} and \emph{same-document, opposite-label negatives}.

\subsection{Paraphrastic Views and Pair Construction}
\label{sec:views}

\paragraph{Label-Preserving Paraphrases}
For each labeled example $(d_i,c_i,y_i)\in\mathcal{D}_s$, CPIL creates alternative surface forms of the document-claim input.
Let $\mathcal{P}$ denote a set of paraphrasing transformations.
A paraphraser $p\in\mathcal{P}$ returns a view \((d_i^{(p)},c_i^{(p)})\).
We define the original view as
\((d_i^{(0)},c_i^{(0)})=(d_i,c_i)\),
and let $\mathcal{V}=\{0\}\cup\mathcal{P}$ be the full view set.

CPIL is based on the \emph{label-preservation condition}: 
if a paraphrastic view preserves the factual content of both the document and the claim, their support relation should also remain unchanged.
Accordingly, each retained view \((d_i^{(p)},c_i^{(p)})\) inherits the original label \(y_i\).

In this work, we instantiate the paraphraser with back-translation \cite{sennrich2016improving,edunov2018understanding}. 
The input document-claim pair is translated into a pivot language and then translated back to the source language.
Back-translation provides surface variation without human rewriting or prompt engineering and can be implemented with off-the-shelf translation systems.
CPIL is not restricted to back-translation; 
other transformations, such as controlled rewriting or QA-style reformulation \cite{yang2025metaqa}, could be used as long as the resulting views provide semantically preserving paraphrases.

\paragraph{Cross-Paraphrase Positive Pairs}
Positive candidates are built from different views of the same labeled example.
Given the view set $\mathcal{V}$, CPIL links distinct views of each document-claim instance:
\begin{equation}
\begin{aligned}
\label{eq:positive_pairs}
\mathcal{D}^{+}
=
\Bigl\{
\bigl((d_i^{(p)},c_i^{(p)}),(d_i^{(q)},c_i^{(q)})\bigr)
& 
:
(d_i,c_i,y_i)\in\mathcal{D}_s,\; 
\\
& \qquad p,q\in\mathcal{V},\;
p<q
\Bigr\}.
\end{aligned}
\end{equation}
Here $p<q$ denotes a fixed ordering over views and avoids duplicate unordered pairs. 
With $|\mathcal{V}|$ views, each labeled example contributes at most $\binom{|\mathcal{V}|}{2}$ positive pairs before downsampling.

These positive pairs provide relational supervision beyond ordinary data augmentation.
Appending paraphrases to the classifier training set exposes the model to more surface forms, but it does not explicitly require their representations to agree.
By contrast, optimization over $\mathcal{D}^+$ explicitly encourages representations of two views of the same document-claim instance to be close. 
This encourages the encoder to focus less on superficial wording and more on the evidence-claim relation that is stable across faithful paraphrases.

\paragraph{Same-document Negative Pairs}

Negative pairs are designed to be more challenging for document-grounded factuality detection. 
The original negative set contains claim pairs that share the same source document but have opposite labels:
\begin{equation}
\label{eq:base_negative_pairs}
\Bigl\{
\bigl((d_i,c_i),(d_j,c_j)\bigr)
:
d_i=d_j,\;
c_i\neq c_j,\;
y_i\neq y_j
\Bigr\}.
\end{equation}
The condition $d_i=d_j$ fixes the evidence source 
and reduces direct cross-document topic and domain shortcuts.
The model must instead distinguish claims with different factuality
labels under the same source document. 

CPIL further allows paraphrastic views on either side of the negative pair:
\begin{equation}
\begin{aligned}
\label{eq:view_negative_pairs}
\mathcal{D}^{-}
=
\Bigl\{
\bigl((d_i^{(p)},c_i^{(p)}),(d_j^{(q)},c_j^{(q)})\bigr)
&
: 
d_i^{(0)}=d_j^{(0)},\;
c_i^{(0)}\neq c_j^{(0)},\; 
\\
& 
y_i\neq y_j,\;
p,q\in\mathcal{V}
\Bigr\}.
\end{aligned}
\end{equation}
The same-document constraint is evaluated using the original source document, because paraphrasing may change surface wording.
These cross-view negatives combine two types of supervision:
they retain the evidence-controlled contrast of Eq.~\eqref{eq:base_negative_pairs}, while introducing additional paraphrastic variability.

Algorithm~\ref{alg:candidate_construction} summarizes the offline construction procedure.
It first generates the available paraphrastic views for each labeled document-claim example, and then constructs the positive and negative candidate pools defined in Eqs.~\eqref{eq:positive_pairs} and~\eqref{eq:view_negative_pairs}. 
This procedure is performed once before model optimization.
The next subsection describes how CPIL dynamically refreshes the negative set during training.

\begin{algorithm}[t]
\caption{Offline Candidate Pool Construction}
\label{alg:candidate_construction}
\begin{algorithmic}[1]
\Require
Labeled subset
\(\mathcal D_s=\{(d_i,c_i,y_i)\}_{i=1}^{N_s}\);
pivot-language set \(\mathcal P\);
back-translation operator \(\operatorname{BT}_p(\cdot)\)
\Ensure
positive and negative candidate pools \(\mathcal D^{+}\), \(\mathcal D^{-}\)

\Statex \textit{View generation}

\State
\(\mathcal V \gets \{0\}\cup\mathcal P\),
where \(0\) denotes the original view

\ForAll{\((d_i,c_i,y_i)\in\mathcal D_s\)}
    \State
    \(d_i^{(0)}\gets d_i\),
    \(c_i^{(0)}\gets c_i\)
    \ForAll{\(p\in\mathcal P\)}
        \State Generate $(d_i^{(p)},c_i^{(p)})$ by back-translation
    \EndFor
\EndFor

\Statex \textit{Candidate-pool construction}

\State Build candidate pools $\mathcal{D}^{-}$ and $\mathcal{D}^+$ using Eqs.~\eqref{eq:positive_pairs} and \eqref{eq:view_negative_pairs}
\State
\Return
\(\mathcal D^{+},\mathcal D^{-}\)
\end{algorithmic}
\end{algorithm}

\subsection{Quantile-Aware Semi-Hard Negative Mining}
\label{sec:semi_hard_mining}
Uniformly sampled same-document negatives are already stronger than random cross-document negatives because they share the same evidence source.
However, many negatives may become easy as training progresses.
Selecting only the most similar negatives is also undesirable:
an extremely similar opposite-label pair may be an informative near-miss, but it may instead reflect paraphrase drift, annotation ambiguity, or insufficient evidence in the source document.

CPIL therefore dynamically selects semi-hard negatives while filtering candidates that become excessively similar to positive pairs.
At the $t$-th training epoch, CPIL first uniformly samples a positive subset with $K$ samples:
$\mathcal{D}^{+}_{K}\subseteq\mathcal{D}^{+}$, where $|\mathcal{D}^{+}_{K}|=K$.
Let $f_{\theta,t}$ be the current encoder. 
For a pair
$r =
    \bigl(
        (d_a,c_a),
        (d_b,c_b)
    \bigr),
$
its mining similarity is
\begin{equation}
    s_t(r, f_{\theta,t})
    =
    \cos\!\left(
        f_{\theta,t}(d_a,c_a),
        f_{\theta,t}(d_b,c_b)
    \right).
    \label{eq:semi_hard_score}
\end{equation}
$f_{\theta,t}$ is used for similarity calculation with gradients disabled.

To avoid selecting negatives that are nearly indistinguishable from positive pairs, CPIL computes a positive-derived safety boundary based on the positive set $\mathcal{D}_K^+$:
\begin{equation}
\tau_t
=
Q_{\rho}
\left(
\left\{
s_t(r):
r\in{\mathcal D}^{+}_{K}
\right\}
\right),
\label{eq:quantile_boundary}
\end{equation}
where $Q_{\rho}$ is the $\rho$-th percentile of positive-pair similarities.
Negative candidates whose similarity exceeds this boundary are treated as overly positive-like and are excluded:
\begin{equation}
\label{eq:safe_pool}
\mathcal{M}^{\mathrm{safe}}_t
=
\{r\in\mathcal{D}^{-}: s_t(r)<\tau_t\}.
\end{equation}

Because the number of safe negatives can vary across epochs, the effective pair budget is
\begin{equation}
    K_t
    =
    \min
    \left\{
        K,
        |\mathcal M_t^{\mathrm{safe}}|
    \right\}.
    \label{eq:effective_pair_budget}
\end{equation}

Within the safe pool, CPIL selects a mixture of difficult and diverse negatives. 
Let $\operatorname{Top}_{m}(\mathcal{A};s_t)$ denote the $m$ elements in a set $\mathcal{A}$ with the largest similarity scores.
The exploitation semi-hard subset is
\begin{equation}
\label{eq:hard_subset}
\mathcal{M}^{\mathrm{hard}}_t
=
\operatorname{Top}_{\lfloor\lambda K_t\rfloor}
(\mathcal{M}^{\mathrm{safe}}_t;s_t),
\end{equation}
where $\lambda\in[0,1]$ controls the fraction of mined semi-hard negatives.
The remaining negatives are uniformly sampled from the rest of the safe pool:
\begin{equation}
\label{eq:coverage_subset}
\mathcal{M}^{\mathrm{cov}}_t
\sim
\operatorname{Uniform}
\left(
\mathcal{M}^{\mathrm{safe}}_t\setminus \mathcal{M}^{\mathrm{hard}}_t,\;
K_t-\lfloor\lambda K_t\rfloor
\right).
\end{equation}
The refreshed negative set is then
\begin{equation}
\label{eq:refreshed_negative}
\mathcal{D}^{-}_{t}
=
\mathcal{M}^{\mathrm{hard}}_t
\cup
\mathcal{M}^{\mathrm{cov}}_t.
\end{equation}


\revisedtext{Mining changes the empirical distribution of negatives used in the ranking objective.}
The boundary conservatively removes negative candidates that fall inside the positive-like region, whereas the exploitation-coverage mixture prevents training from repeatedly concentrating on a small set of highly similar candidates.

Equations~\eqref{eq:semi_hard_score}-\eqref{eq:refreshed_negative} define the epoch-wise negative-selection procedure.
Algorithm~\ref{alg:dynamic_training} integrates this procedure with the two-stage CPIL network training in Sec. \ref{sec:training}. 

\begin{algorithm}[t]
\caption{Dynamic Mining and Two-Stage Training}
\label{alg:dynamic_training}
\begin{algorithmic}[1]
\Require
Labeled subset \(\mathcal D_s\);
candidate pools \(\mathcal D^{+},\mathcal D^{-}\) from
Algorithm~\ref{alg:candidate_construction};
encoder \(f_\theta\);
classifier \(g_\phi\);
Stage~1 epochs \(T_1\);
Stage~2 epochs \(T_2\);
pair-budget upper bound \(K\);
quantile level \(\rho\);
mining fraction \(\lambda\);
ranking margin \(\alpha\)

\Ensure
Trained detector
\(h(d,c)=\sigma(g_\phi(f_\theta(d,c)))\)

\Statex \textit{Stage~1: Pairwise Encoder Training}

\For{\(t=1,\ldots,T_1\)}
    \State 
    Sample $\mathcal{D}^+_K \sim \mathrm{Uniform}(\mathcal{D^+},K)$
    \State
    Compute \(s_t(r)\) for all \(r\in{\mathcal D}^{+}_{K} \cup \mathcal{D}^-\)
    using Eq. (\ref{eq:semi_hard_score})
    \State
    Compute \(\tau_t\) using Eq.~\eqref{eq:quantile_boundary}
    \State
    Construct
    \(\mathcal M_t^{\mathrm{safe}}\) using Eq.~\eqref{eq:safe_pool}
    \State
    Compute \(K_t\) using Eq.~\eqref{eq:effective_pair_budget}

    \If{\(K_t>0\)}
        \State
        Select \(\mathcal M_t^{\mathrm{hard}}\) using Eq.~\eqref{eq:hard_subset}
        \State
        Sample \(\mathcal M_t^{\mathrm{cov}}\) using Eq.~\eqref{eq:coverage_subset}
        \State
        Set \(\mathcal D^{-}_{t} \gets \mathcal M_t^{\mathrm{hard}} \cup \mathcal M_t^{\mathrm{cov}}\)
        \State
        Sample $\mathcal{D}^+_t \sim \mathrm{Uniform}(\mathcal{D}_K^+,K_t)$
        \State
        Update \(\theta\) using \(\mathcal L_{\mathrm{ctr}}\) in Eq.~\eqref{eq:ctr}
    \EndIf
\EndFor

\Statex \textit{Stage~2: Classifier Fine-Tuning}

\State
Set $f_\theta \leftarrow f_{\theta,T_1}$ and initialize \(g_\phi\)
\For{\(t=1,\ldots,T_2\)}
    \State
    Jointly update \(\theta,\phi\) on \(\mathcal D_s\) using \(\mathcal L_{\mathrm{ft}}\) in Eq. (\ref{eq:stage2_loss})
\EndFor

\State
\Return
\(h(d,c)=\sigma(g_\phi(f_\theta(d,c)))\)
\end{algorithmic}
\end{algorithm}

\subsection{CPIL Network Training}
\label{sec:training}

CPIL consists of a shared encoder \(f_\theta\) and a linear classification head \(g_\phi\). 
Given a packed document-claim input \((d,c)\), the encoder produces the embedding
\begin{equation}
    \mathbf e
    =
    f_\theta(d,c).
\end{equation}
The encoder can be instantiated with a pretrained Transformer such
as DeBERTa~\cite{he2021debertav3}.
In Stage~2, the linear classification head produces a scalar factuality logit
\begin{equation}
    g_\phi(\mathbf e)
    =
    \mathbf w_\phi^\top\mathbf e+b_\phi.
    \label{eq:linear_head}
\end{equation}

\paragraph{Stage~1: Siamese Pair-Similarity Pretraining}

The first stage trains the encoder before attaching the final classifier. 
Two weight-tied towers of $f_{\theta,t}$ process paired inputs drawn from $\mathcal{D}^+_t \bigcup \mathcal{D}_t^-$. 
Its objective (i) pulls together cross-paraphrase views of the same instances in $\mathcal{D}^+_t$ and (ii) repels same-document pairs with opposite labels in $\mathcal{D}^-_t$:
\begin{equation}
\mathcal L_{\mathrm{ctr}}
=
\frac{1}{K_t}
\sum_{k=1}^{K_t}
\max
\left\{
0,\,
s_t(r_{k}^{-})
-
s_t(r_{k}^{+})
+
\alpha
\right\},
\label{eq:ctr}
\end{equation}
where $\alpha>0$ is the margin,
and $s(r)$ indicates the similarity of two document-claim pairs as defined in Eq.~\eqref{eq:semi_hard_score}, but with gradients enabled during optimization.
Minimizing Eq.~\eqref{eq:ctr} encourages cross-paraphrase positives to be closer than same-document opposite-label negatives by at least $\alpha$.

Other contrastive objectives, such as NT-Xent \cite{chen2020simclr} or supervised contrastive learning \cite{khosla2020contrastive_learning,gunel2020supervised}, could be used with the same candidate construction.
CPIL's primary contribution lies in its factuality-aware pair definitions and dynamic negative selection rather than a particular contrastive loss.

\paragraph{Stage~2: Classifier Fine-Tuning}
After Stage~1, CPIL attaches a classification head in Eq. (\ref{eq:linear_head}).
The detector predicts 
\[
h(d,c)=\sigma(g_{\phi}(f_{\theta}(d,c))),
\]
where $\sigma(\cdot)$ is the sigmoid function.
The encoder is initialized from Stage~1 and then fine-tuned together with the classifier on the original labeled subset $\mathcal{D}_s$:
\begin{equation}
\label{eq:stage2_loss}
\mathcal{L}_{\mathrm{ft}}
=
\mathbb{E}_{(d,c,y)\sim\mathcal{D}_s}
\left[
\operatorname{BCE}
\bigl(
\sigma(g_{\phi}(f_{\theta}(d,c))),y
\bigr)
\right],
\end{equation}
where $\operatorname{BCE}$ denotes binary cross-entropy.
Stage~2 converts the representation learned from pairwise constraints into a binary factuality predictor.

Paraphrase generation, pair construction, and dynamic mining are used only during training.
At inference time, CPIL uses the standard encoder-classifier pipeline without Siamese comparison or paraphrase generation.

\begin{table*}[h]
\centering
\caption{
Task-level F1 (\%) and the unweighted average (AVG) over 11 LLM-AggreFact tasks.
The best and second-best results are shown in bold and underlined, respectively.
The CPIL-3\% row reports the mean over three random seeds, and the STD row gives the corresponding standard deviation.
}
\label{tab:main_result}
\resizebox{\textwidth}{!}{
\begin{tabular}{l|ccccccccccc|c}
\toprule
\multirow{2}{*}{Model}
& \multicolumn{2}{c}{AggreFact}
& \multicolumn{2}{c}{TofuEval}
& \multirow{2}{*}{WiCE}
& \multirow{2}{*}{REVEAL}
& \multicolumn{1}{c}{Claim}
& \multicolumn{1}{c}{Fact}
& \multicolumn{1}{c}{Expert}
& \multirow{2}{*}{LFQA}
& \multicolumn{1}{c|}{RAG}
& \multirow{2}{*}{AVG}
\\
& CNN & XSum & MediaS & MeetB &  &  & Verify & Check & QA &  & Truth &  \\
\midrule
Qwen2.5-7B 
& 92.41 & 71.25 & 86.21 & \ru{90.16} & 53.47 & 74.09 & 82.90 & 61.28 & 75.84 & 83.40 & \ru{89.18} & 78.20 \\ 
Llama3.1-8B 
& 92.65 & 71.82 & 86.21 & \win{91.43} & 57.42 & 69.79 & 83.79 & 59.58 & 84.42 & 81.04 & \win{94.02} & 79.29 \\ 
Mistral-Nemo-12B & \ru{94.07} & 72.14 & 85.82 & 74.28 & 60.53 & \win{84.43} & 88.04 & \win{77.47} & 85.00 & 86.40 & 57.38 & 78.69 \\
\midrule 
SummaC-ZS     
& 55.19 & 66.55 & 79.57 & 80.14 & 41.03 & 74.50 & 72.96 & 60.79 & 65.30 & 81.58 & 62.81 & 67.31 \\
SummaC-CV     
& 66.41 & 48.56 & 69.70 & 77.87 & 50.98 & 70.77 & 68.36 & 57.53 & 65.21 & 79.16 & 56.92 & 64.68 \\
AlignScore    
& 81.71 & 68.51 & 80.59 & 85.18 & 54.75 & 69.79 & 81.16 & 61.05 & 70.63 & 85.81 & 77.35 & 74.23 \\
MiniCheck     
& 74.88 & \ru{74.49} & 83.72 & 81.26 & 62.10 & 72.31 & 85.54 & 60.15 & 72.22 & 88.15 & 86.36 & 76.47 \\
FactCG        
& 78.44 & 74.46 & 83.98 & 80.07 & \ru{69.87} & 80.53 & 83.45 & 63.42 & 58.54 & 89.12 & 86.63 & 77.14 \\
CCHD 
& 88.05 & \win{77.10} & 84.74 & 87.43 & \win{70.70} & 78.29 & 83.81 & 61.47 & 68.53 & 89.46 & 87.40 & 79.73 \\
\midrule
CPIL-2\% 
& 93.80 & 70.15 & \ru{86.85} & 88.05 & 56.39 & 81.88 & \ru{89.96} & 65.61 & \ru{86.35} & \win{90.04} & 82.40 & \ru{81.04} \\
CPIL-3\% 
& \win{94.77} & 72.34 & \win{87.40} & 87.05 & 55.80 & \ru{83.48} & \win{90.01} & \ru{68.41} & \win{87.36} & \ru{89.62} & 81.33 & \win{81.60} \\
STD & $\pm$0.15 & $\pm$0.27 & $\pm$0.51 & $\pm$0.03 & $\pm$1.20 & $\pm$1.11 & $\pm$0.03 & $\pm$1.85 & $\pm$0.57 & $\pm$0.42 & $\pm$0.66 & $\pm$0.19 \\

\bottomrule
\end{tabular}
}
\end{table*}

\subsection{Algorithmic Summary and Computational Complexity}
\label{sec:complexity}

Algorithms~\ref{alg:candidate_construction} and~\ref{alg:dynamic_training} separate CPIL into an \textit{offline} construction stage and an \textit{online} optimization stage.
Algorithm~\ref{alg:candidate_construction} generates the paraphrastic views and constructs the fixed candidate pools.
Algorithm~\ref{alg:dynamic_training} then reuses these pools during Stage~1, refreshing the negatives at every epoch before fine-tuning the factuality classifier in Stage~2.
Unless otherwise stated, all sampling operations are performed without replacement.

Let $|\mathcal D^{+}|$ and $|\mathcal D^{-}|$ denote the positive and negative candidate-pool sizes, respectively.
Algorithm~\ref{alg:candidate_construction} is executed once before training.
Excluding the external cost of back-translation, enumerating and materializing the candidate-pair indices requires $O(|\mathcal D^{+}|+|\mathcal D^{-}|)$ time and storage.

At each Stage~1 epoch,
Algorithm~\ref{alg:dynamic_training} scores $K$ sampled positive pairs and $|\mathcal D^{-}|$ negative candidates.
The pair-level scoring and selection operations therefore scale as $O(K+|\mathcal D^{-}|)$,
in addition to the encoder computation required to obtain their representations.
A straightforward implementation may encode the same view multiple times when it occurs in different pairs. 
Caching the embeddings of shared views can reduce this repeated computation to the number of distinct views appearing in the scored candidate set.

After mining, the ranking update uses $K_t$ matched positive-negative training units in epoch $t$. 
Stage~2 has the same asymptotic training cost as standard fine-tuning on the labeled subset $\mathcal D_s$.
Because paraphrase generation, candidate construction, and dynamic mining are restricted to training, CPIL introduces no pairwise or mining operations at inference time.
Its deployment cost is therefore that of the underlying encoder-classifier under the same input-processing protocol.

\section{Experiments}

\subsection{Experiment Settings}
\label{sec:experiemnt_settings}

\paragraph{Benchmark and Evaluation Metrics}
We evaluate CPIL on \textit{LLM-AggreFact}~\cite{tang2024minicheck}, an aggregate benchmark consisting of 11 document-grounded factuality tasks. 
The benchmark covers summarization, dialogue, claim verification, long-form question answering, and retrieval-augmented generation.
The official development and test splits contain 30,420 and 29,320 examples, respectively.
We report the F1 score (\%) on each task and the macro-F1 over the
11 tasks, denoted by AVG. 
Macro averaging assigns equal weight to each task and therefore avoids allowing a large constituent dataset to dominate the overall result.

\paragraph{Baselines}
We compare CPIL with representative non-LLM factuality detectors: \textit{SummaC-ZS} and \textit{SummaC-CV}~\cite{laban2022summac}, \textit{AlignScore}~\cite{zha2023alignscore}, \textit{MiniCheck}~\cite{tang2024minicheck}, \textit{FactCG}~\cite{lei2025factcg}, and \textit{CCHD}~\cite{lin2026cchd}. 
These baselines cover NLI-based consistency scoring, learned alignment functions, efficient document-grounded verification, graph-based synthetic claim augmentation, and paraphrase-consistency-constrained training.
When available, we use authors' released implementations and recommended checkpoints. 
We additionally report Qwen2.5-7B~\cite{qwen2.5}, Llama3.1-8B~\cite{llama3}, \revisedtext{and Mistral-Nemo-12B}~\cite{mistralnemo2024} as LLM references.

\paragraph{Implementation Details}
CPIL instantiates the shared encoder \(f_\theta\) with DeBERTa-v3-large~\cite{he2021debertav3} and uses a linear classification head \(g_\phi\).
Each document-claim example is encoded as a paired input sequence.
The maximum input length is 2,046 tokens for training.


We generate back-translated views (using Google Translate API) via three pivot languages, 
$\mathcal P = \{\texttt{fr},\texttt{es},\texttt{zh}\}$,
corresponding to French, Spanish, and Chinese.
The document and claim are translated separately through the same pivot language.
Together with the original input, each example can therefore have up to four views.
\revisedtext{The training pipeline consumes cached back-translated views and retains every serialized view without manual screening. 
The similarity-based negative filter in Sec. III-C operates after pair construction.}

The labeled training pool follows the data released with FactCG~\cite{lei2025factcg} and contains approximately 16,700 labeled document-claim examples. 
Applying the three-pivot view-generation and pair-construction procedure to the full pool yields more than 350,000 candidate pairs.
To evaluate CPIL under restricted supervision, we use either 2\% or 3\% of this labeled pool, yielding approximately 7,000 and 12,000 candidate training pairs, respectively. 
We denote the resulting models by \textit{CPIL-2\%} and
\textit{CPIL-3\%}, respectively. 
\revisedtext{
A source document with only one factuality class in $\mathcal D_s$ contributes no same-document negative pairs. 
$\mathcal D_s^{2\%}\subseteq\mathcal D_s^{3\%}$ holds.
}

For Stage~1, the pair budget is set separately for positive and negative pairs.
We set the ranking margin \(\alpha=1.5\), and the semi-hard proportion \(\lambda=0.8\).
\revisedtext{The lower positive-similarity quantile \(\rho=10\%\) heuristically. 
With $\rho$ fixed, the numerical boundary $\tau_t$ is recomputed from the current positive-pair similarities at each Stage 1 epoch.}
\(K\)=1,500 per pair type for CPIL-2\%, and \(K\)=2,000 per pair type for CPIL-3\%.
We optimize CPIL using AdamW with a learning rate of \(1\times10^{-5}\), an Adam epsilon of \(1\times10^{-6}\), zero weight decay, and a linear warmup ratio of 0.06.
The training batch size is 2, with gradient accumulation over 8 batches, resulting in an effective batch size of 16.
We train Stage~1 for at most three epochs and Stage~2 for one epoch.

\paragraph{Threshold Protocol}
For each task, we first score its development examples, and choose the threshold that maximizes supported-class development F1. 
If multiple thresholds tie, we first select the one closest to 0.5 and then the larger threshold. 
The selected threshold is then frozen and applied to the corresponding test task. 
The same threshold-selection procedure is applied to all methods. 

\subsection{Main Results}
\label{sec:main_results}

Table~\ref{tab:main_result} reports the task-level and macro-F1 results on LLM-AggreFact.
CPIL achieves the highest macro-F1 among the compared compact factuality detectors under both low-label settings.
CPIL-2\% obtains an AVG of 81.04 using 2\% of the FactCG-released labeled pool, outperforming CCHD by 1.31 points and FactCG by 3.90 points.
Increasing the label budget to 3\% further raises the AVG to 81.60,
and CPIL-3\% also exceeds the two reported vanilla LLM references in macro-F1. 
These results show that the available low-budget labels can provide strong supervision when they are converted into task-aware pairwise constraints.


\begin{table*}[t]
\centering
\caption{
%
Ablation results under the 2\% label budget. 
If available, 
all variants use the same labeled subset, encoder, pair budget, and optimization settings unless otherwise specified.
} 
\label{tab:ablation_study}
\resizebox{\textwidth}{!}{
\begin{tabular}{c|ccccccccccc|c}
\toprule
\multirow{2}{*}{Variant}
& \multicolumn{2}{c}{AggreFact}
& \multicolumn{2}{c}{TofuEval}
& \multirow{2}{*}{WiCE}
& \multirow{2}{*}{REVEAL}
& \multicolumn{1}{c}{Claim}
& \multicolumn{1}{c}{Fact}
& \multicolumn{1}{c}{Expert}
& \multirow{2}{*}{LFQA}
& \multicolumn{1}{c|}{RAG}
& \multirow{2}{*}{AVG}
\\
& CNN & XSum & MediaS & MeetB &  &  & Verify & Check & QA &  & Truth &  \\
\midrule
\textit{w/o} Pairwise Stage 
& 70.94 & 69.92 & 81.88 & 88.52 & 68.46 & 76.56 & 83.02 & 60.47 & 72.25 & 89.24 & 83.58 & 76.81 \\
Append-Only Aug. 
& 78.30 & 69.20 & 82.46 & 86.74 & 69.90 & 77.77 & 86.41 & 61.28 & 65.15 & 87.47 & 86.10 & 77.34 \\
Random All Pairs 
& 77.16 & 74.15 & 79.84 & 83.52 & 70.31 & 76.49 & 85.04 & 61.69 & 70.70 & 87.69 & 85.33 & 77.44 \\
Random Pos Pairs
& 94.61 & 69.45 & 84.38 & 89.03 & 47.02 & 85.13 & 95.97 & 78.40 & 86.56 & 89.16 & 46.74 & 78.77 \\
Random Neg Pairs 
& 94.71 & 69.14 & 84.52 & 88.98 & 49.22 & 74.70 & 95.96 & 57.26 & 86.56 & 89.24 & 47.34 & 76.15 \\
\textit{w/o} Dynamic Mining 
& 91.08 & 71.38 & 84.05 & 90.10 & 64.91 & 77.77 & 81.74 & 58.36 & 71.94 & 88.90 & 84.79 & 78.64 \\
\textbf{Full CPIL-2\%} & 93.80 & 70.15 & 86.85 & 88.05 & 56.39 & 81.88 & 89.96 & 65.61 & 86.35 & 90.04 & 82.40 & 81.04 \\
\bottomrule
\end{tabular}
}
\end{table*}

\subsection{Ablation Study}
\label{sec:ablation}
Table~\ref{tab:ablation_study} evaluates the contribution of the pairwise pretraining stage, the factuality-aware pair definitions, and dynamic negative mining. 
We compare Full CPIL with six variants.

\begin{itemize}
    \item \textit{w/o Pairwise Stage} removes Stage~1 and directly trains a classifier on the original labeled examples.
    \item \textit{Append-Only Aug.} also removes Stage~1, but appends paraphrased views to the classifier training set as ordinary augmented samples.
    \item \textit{Random All Pairs} keeps Stage~1 but randomly constructs both positive and negative pairs.
    \item \textit{Random Pos Pairs} replaces cross-paraphrase positives with randomly matched positive pairs.
    \item \textit{Random Neg Pairs} replaces same-document opposite-label negatives with randomly matched negative pairs.
    \item \textit{w/o Dynamic Mining} retains the two-stage framework and the proposed pair definitions but replaces quantile-aware semi-hard mining with static negative sampling.
\end{itemize}

\paragraph{Effect of Pairwise Pretraining}
Removing Stage~1 decreases AVG from 81.04 to 76.81, a reduction of 4.23 points.
\textit{Append-only} augmentation raises AVG from 76.81 to 77.34 but remains 3.70 points below Full CPIL.
Thus, merely increasing the number of paraphrastic training examples does not reproduce the gain obtained by explicitly aligning different views through pairwise pretraining.

\paragraph{Effect of Factuality-Aware Pair Construction}
\textit{Random All Pairs} obtains an AVG of 77.44, which is close to \textit{Append-Only Aug.}
This result indicates that the benefit of Stage~1 depends on how the pairs are defined rather than on pairwise training alone.
Generic random pairs provide weak or noisy supervision for document-grounded factuality.
\textit{Random Pos Pairs} reaches an AVG of 78.77 but produces highly uneven task-level results, with severe drops on WiCE and RAGTruth.
The result supports constructing positives from different paraphrastic views of the same document-claim example rather than from arbitrary instances that only share a label.
\textit{Random Neg Pairs} drops to 76.15 AVG, showing that negative pairs must preserve the evidence context. 
Without the same-document constraint, many negatives can be separated by topic or document mismatch rather than by factual support.

\revisedtext{Taken together, Append-Only Aug. (77.34), Random All Pairs (77.44), and Full CPIL (81.04) support the importance of task-aware relational structure, rather than merely more augmented examples or a pairwise training stage.}

\paragraph{Effect of Dynamic Negative Mining}
Replacing dynamic mining with static negative sampling reduces AVG to 78.64.
Because this variant retains the two-stage architecture and the same positive and negative candidate definitions, the 2.40-point difference provides evidence for the contribution of adaptive negative selection under the tested configuration.
Static same-document negatives are useful, but they may become easy as the encoder improves.
Full CPIL refreshes the negative set according to the current representation space, selects informative semi-hard negatives, and retains coverage through uniform sampling from the safe pool.

\begin{figure*}
    \centering
    \subfloat[Sensitivity to labeled data size.]{
    \includegraphics[width=0.48\linewidth]{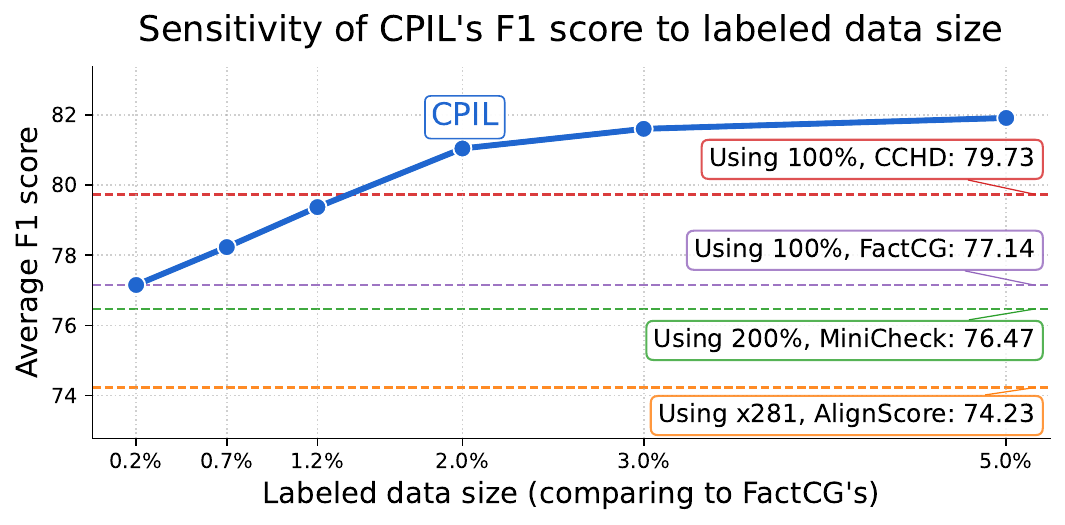}
    \label{fig:sensitivity_label}
    }
    \hfill
    \subfloat[Sensitivity to semi-hard ratio $\lambda$.]{
        \includegraphics[width=0.48\linewidth]{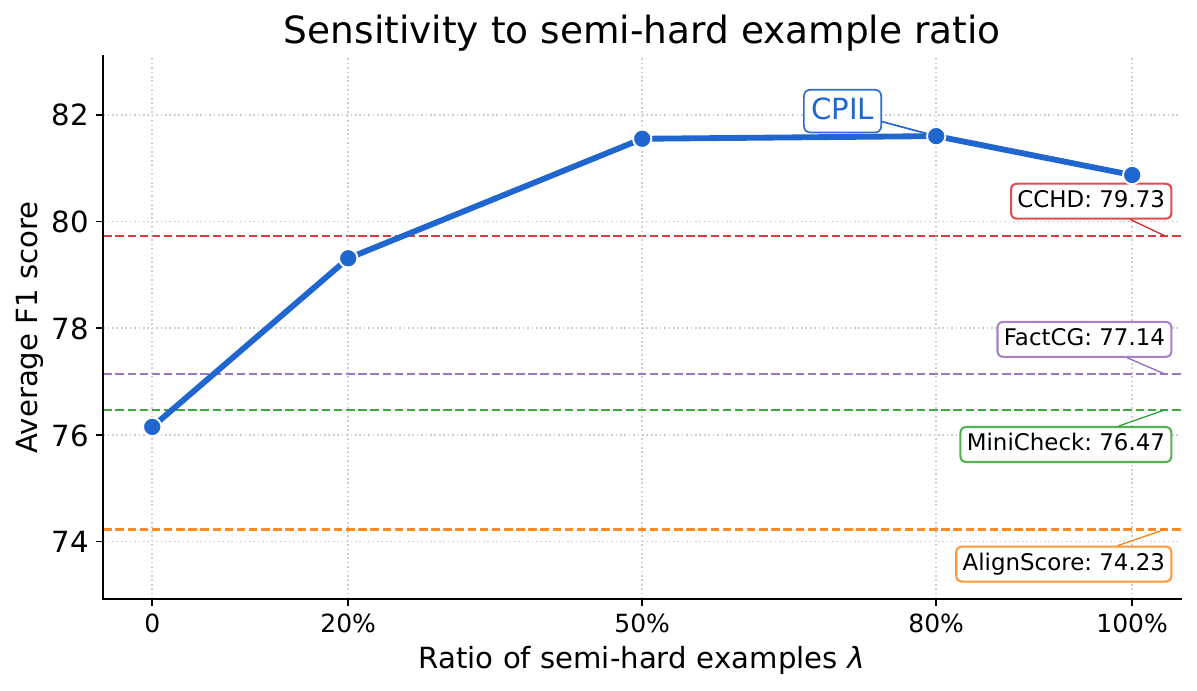}
        \label{fig:sensitivity_semihard}
    }
    \caption{
    Sensitivity analysis of CPIL.
    Left: macro-F1 under different labeled data budgets.
    Right: macro-F1 under different proportions \(\lambda\) of semi-hard negatives in the selected negative set. 
    Performance improves rapidly in the lowest-label regimes and remains
    strong over a range of mixed semi-hard configurations. 
    }
    \label{fig:sensitivity}
\end{figure*}

\begin{figure*}
    \centering
    \subfloat[2\% label budget.]{
        \includegraphics[width=0.45\linewidth]{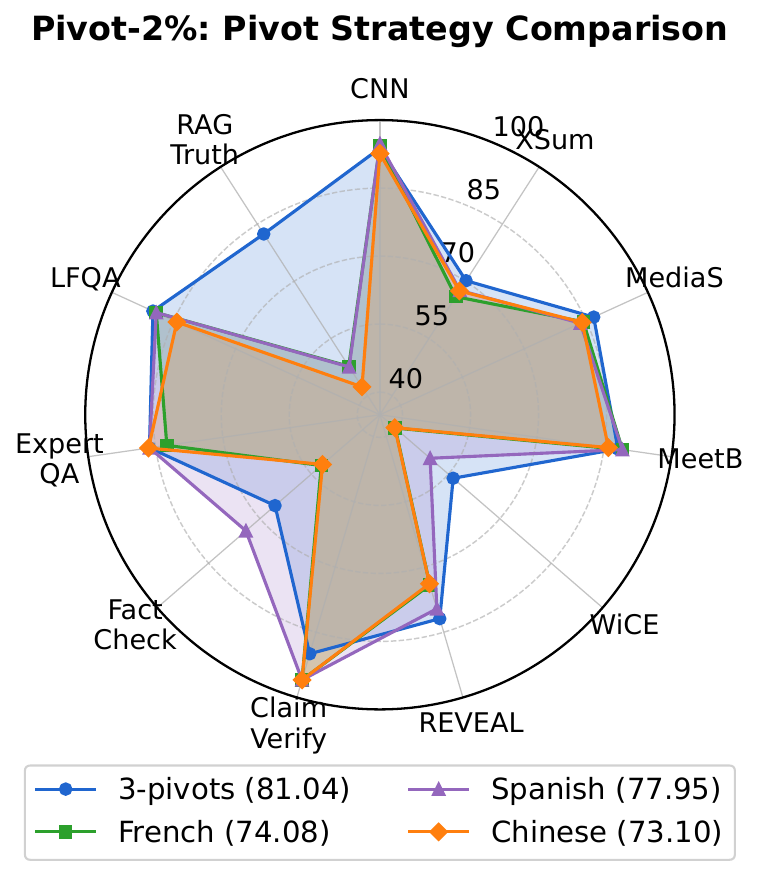}
        \label{fig:pivot_2}
    }
    \hfill
    \subfloat[3\% label budget.]{
        \includegraphics[width=0.45\linewidth]{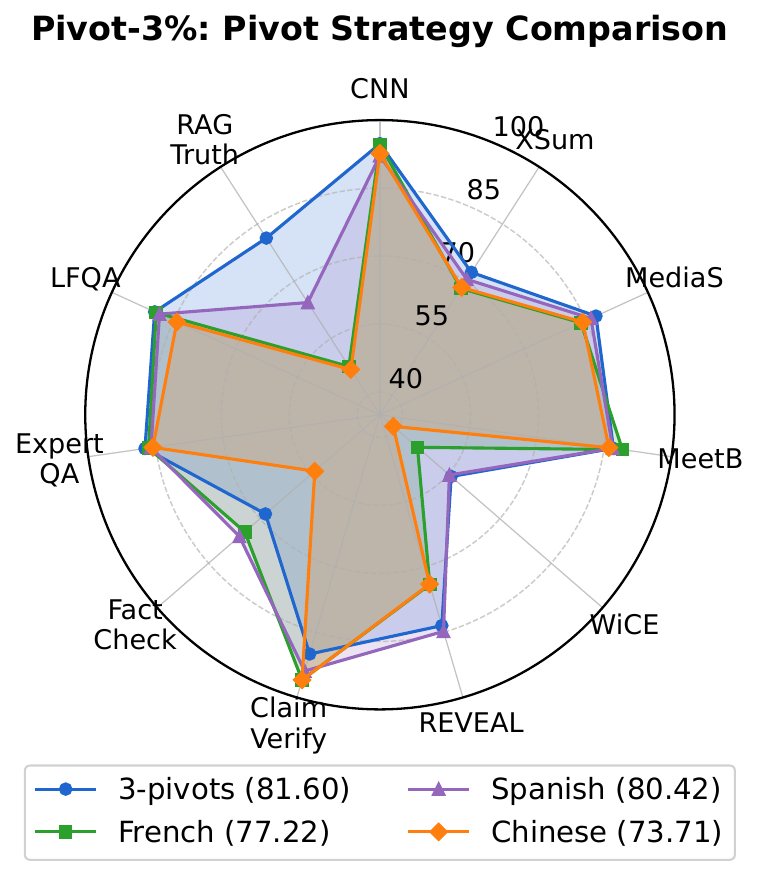}
        \label{fig:pivot_3}
    }
    \caption{
    Macro-F1 comparison of single- and three-pivot back-translation settings under the 2\% and 3\% label budgets. 
    \revisedtext{Each radar axis corresponds to one task; values in parentheses in the legends are the unweighted mean over all 11 tasks (macro-F1/AVG).}
    }
    \label{fig:pivot}
\end{figure*}

\subsection{Sensitivity Analysis}
\label{sec:sensitivity}

We analyze CPIL with respect to the labeled-data fraction and the
semi-hard negative proportion \(\lambda\).

\paragraph{Labeled-Data Fraction}

Fig.~\ref{fig:sensitivity_label} varies the labeled subset from 0.2\% to 5.0\% of the FactCG-released labeled pool.
Macro-F1 generally increases as more labeled examples become available, with the largest gains occurring in the lowest-budget range.
At 0.2\%, CPIL already reaches a macro-F1 level comparable to several compact baselines.
The curve continues to improve through the 2\% and 3\% settings and changes only modestly between 3\% and 5\%.

This pattern is consistent with CPIL's relational-supervision design.
At very low budgets, each additional labeled example introduces not only one classification target but also multiple candidate views and pairwise relations.
As the labeled pool grows, the marginal benefit becomes smaller, indicating that increasing the number of labeled examples alone does not yield proportional gains beyond the moderate-budget regime.
\revisedtext{However, the label-efficiency claim mainly concerns reuse of selected factuality annotations, but not about training time or total compute, given extra one-time back-translation cost.}

\paragraph{Semi-Hard Negative Proportion}

Fig.~\ref{fig:sensitivity_semihard} varies \(\lambda\), the proportion of semi-hard negatives in the selected negative set.
When \(\lambda=0\), all selected negatives are uniformly sampled from the safe pool, resulting in substantially lower macro-F1 than the mixed-selection settings.
Increasing \(\lambda\) improves performance, with the strongest results appearing in the range from 0.5 to 0.8.
The default setting \(\lambda=0.8\) achieves near-peak performance while assigning most of the pair budget to semi-hard negatives.

When \(\lambda=1\), all selected negatives come from the highest similarity region of the safe pool, and performance decreases slightly relative to the mixed settings.
This result supports retaining a coverage component: 
semi-hard examples provide informative comparisons, whereas uniformly sampled safe negatives prevent the selected set from concentrating exclusively on a narrow group of high-similarity pairs.

\subsection{Effect of Pivot Languages}
\label{sec:pivot}

We examine whether CPIL benefits from combining multiple back-translation paths.
Figs.~\ref{fig:pivot_2} and~\ref{fig:pivot_3} compare the three-pivot configuration with French-, Spanish-, and Chinese-only variants under the 2\% and 3\% label budgets.

Under the 2\% budget, the three-pivot setting obtains an AVG of 81.04, compared with 74.08 for French, 77.95 for Spanish, and 73.10 for Chinese.
Under the 3\% budget, the corresponding results are 81.60, 77.22, 80.42, and 73.71.
Thus, the three-pivot setting achieves the highest macro-F1 under both supervision budgets. 

Spanish is the strongest single-pivot option in both settings, but it remains 3.09 points below the three-pivot configuration under the 2\% budget and 1.18 points below it under the 3\% budget.
The benefit of using multiple pivots is therefore larger in the lower-label setting.
One possible explanation is that additional pivot paths provide more varied pairwise relations when only a small number of labeled anchors are available.

The task-level plots also show that the relative ranking of pivot languages varies across datasets.
No single pivot consistently dominates all tasks.
The three-pivot configuration reduces dependence on one translation path and empirically provides a stronger overall setting for CPIL.

\section{Conclusion}

This paper presented CPIL, a two-stage Siamese framework for label-efficient document-grounded factuality detection.
CPIL converts a limited set of labeled document-claim examples into relational supervision through cross-paraphrase positive pairs and same-document, opposite-label negative pairs.
It further introduces quantile-aware semi-hard negative mining, which filters overly positive-like negative candidates and balances negative difficulty with training-set coverage.

On the 11-task LLM-AggreFact benchmark, CPIL-2\% and CPIL-3\% achieve macro-F1 scores of 81.04 and 81.60, respectively.
Both settings outperform the six compact factuality detectors and two vanilla LLM references.
The ablation results further show that replacing dynamic mining with static negative sampling decreases macro-F1 to 78.64, while append-only augmentation and generic random pairing are also less effective than the proposed task-aware relational supervision.

A \textit{limitation} of CPIL is that back-translated views are assumed to preserve the original factuality label. 
Small changes to entities, quantities, temporal expressions, or negation may alter the document-claim support relation. 
In addition, the current evaluation is limited to an English aggregate benchmark and relies on an external translation service for offline view generation.
Future work will study entailment- or entity-aware filtering, alternative paraphrasers, and broader cross-domain and multilingual evaluation \cite{chen2024unified}.

\section*{Acknowledgments}
This work was supported by the National Science Foundation of China (No. 62476060).
Sihong Xie was supported by the Department of Science and Technology of Guangdong Province (No. 2023CX10X079).

\bibliographystyle{IEEEtran}
\bibliography{main}

@article{rahman2025hallucination_review,
  title={Hallucination to truth: a review of fact-checking and factuality evaluation in large language models},
  author={Rahman, Subhey Sadi and Islam, Md Adnanul and Alam, Md Mahbub and Zeba, Musarrat and Rahman, Md Abdur and Chowa, Sadia Sultana and Raiaan, Mohaimenul Azam Khan and Azam, Sami},
  journal={Artificial Intelligence Review},
  year={2026},
  publisher={Springer}
}

@article{krishna2025safety,
  title={Importing Phantoms: Measuring LLM Package Hallucination Vulnerabilities},
  author={Krishna, Arjun and Galinkin, Erick and Derczynski, Leon and Martin, Jeffrey},
  journal={arXiv preprint arXiv:2501.19012},
  year={2025}
}

@article{laban2022summac,
  title={SummaC: Re-visiting NLI-based models for inconsistency detection in summarization},
  author={Laban, Philippe and Schnabel, Tobias and Bennett, Paul N and Hearst, Marti A},
  journal={Transactions of the Association for Computational Linguistics},
  volume={10},
  pages={163--177},
  year={2022},
  publisher={MIT Press One Rogers Street, Cambridge, MA 02142-1209, USA journals-info~…}
}

@inproceedings{zha2023alignscore,
    title = "{A}lign{S}core: Evaluating Factual Consistency with A Unified Alignment Function",
    author = "Zha, Yuheng  and
      Yang, Yichi  and
      Li, Ruichen  and
      Hu, Zhiting",
    booktitle = "Proceedings of the 61st Annual Meeting of the Association for Computational Linguistics (Volume 1: Long Papers)",
    year = "2023",
    address = "Toronto, Canada",
    publisher = "Association for Computational Linguistics",
    doi = "10.18653/v1/2023.acl-long.634",
    pages = "11328--11348",
}

@inproceedings{tang2024minicheck,
    title = "{M}ini{C}heck: Efficient Fact-Checking of {LLM}s on Grounding Documents",
    author = "Tang, Liyan  and
      Laban, Philippe  and
      Durrett, Greg",
    booktitle = "Proceedings of the 2024 Conference on Empirical Methods in Natural Language Processing",
    year = "2024",
    doi = "10.18653/v1/2024.emnlp-main.499",
    pages = "8818--8847",
}

@inproceedings{manakul2023selfcheckgpt,
  title={Selfcheckgpt: Zero-resource black-box hallucination detection for generative large language models},
  author={Manakul, Potsawee and Liusie, Adian and Gales, Mark},
  booktitle={Proceedings of the 2023 conference on empirical methods in natural language processing},
  pages={9004--9017},
  year={2023}
}

@inproceedings{liu2023geval,
    title = "{G}-Eval: {NLG} Evaluation using Gpt-4 with Better Human Alignment",
    author = "Liu, Yang  and
      Iter, Dan  and
      Xu, Yichong  and
      Wang, Shuohang  and
      Xu, Ruochen  and
      Zhu, Chenguang",
    booktitle = "Proceedings of the 2023 Conference on Empirical Methods in Natural Language Processing",
    year = "2023",
    publisher = "Association for Computational Linguistics",
    doi = "10.18653/v1/2023.emnlp-main.153",
    pages = "2511--2522",
}

@inproceedings{lei2025factcg,
    title = "{F}act{CG}: Enhancing Fact Checkers with Graph-Based Multi-Hop Data",
    author = "Lei, Deren  and
      Li, Yaxi  and
      Li, Siyao  and
      Hu, Mengya  and
      Xu, Rui  and
      Archer, Ken  and
      Wang, Mingyu  and
      Ching, Emily  and
      Deng, Alex",
    booktitle = "Proceedings of the 2025 Conference of the Nations of the Americas Chapter of the Association for Computational Linguistics: Human Language Technologies (Volume 1: Long Papers)",
    year = "2025",
    publisher = "Association for Computational Linguistics",
    doi = "10.18653/v1/2025.naacl-long.258",
    pages = "5002--5020",
    ISBN = "979-8-89176-189-6",
}

@article{muhammed2025selfcheckagent,
  title={SelfCheckAgent: Zero-Resource Hallucination Detection in Generative Large Language Models},
  author={Muhammed, Diyana and Rabby, Gollam and Auer, S{\"o}ren},
  journal={arXiv preprint arXiv:2502.01812},
  year={2025}
}

@inproceedings{kryscinski2019factcc,
    title = "Evaluating the Factual Consistency of Abstractive Text Summarization",
    author = "Kryscinski, Wojciech  and
      McCann, Bryan  and
      Xiong, Caiming  and
      Socher, Richard",
    booktitle = "Proceedings of the 2020 Conference on Empirical Methods in Natural Language Processing (EMNLP)",
    year = "2020",
    publisher = "Association for Computational Linguistics",
    doi = "10.18653/v1/2020.emnlp-main.750",
    pages = "9332--9346",
}

@inproceedings{fabbri2021qafacteval,
    title = "{QAF}act{E}val: Improved {QA}-Based Factual Consistency Evaluation for Summarization",
    author = "Fabbri, Alexander  and
      Wu, Chien-Sheng  and
      Liu, Wenhao  and
      Xiong, Caiming",
    booktitle = "Proceedings of the 2022 Conference of the North American Chapter of the Association for Computational Linguistics: Human Language Technologies",
    year = "2022",
    publisher = "Association for Computational Linguistics",
    doi = "10.18653/v1/2022.naacl-main.187",
    pages = "2587--2601",
}

@article{yang2025metaqa,
  title={Hallucination Detection in Large Language Models with Metamorphic Relations},
  author={Yang, Borui and Al Mamun, Md Afif and Zhang, Jie M and Uddin, Gias},
  journal={Proceedings of the ACM on Software Engineering},
  volume={2},
  number={FSE},
  pages={425--445},
  year={2025},
  publisher={ACM New York, NY, USA}
}

@article{beddiar2021data,
  title={Data expansion using back translation and paraphrasing for hate speech detection},
  author={Beddiar, Djamila Romaissa and Jahan, Md Saroar and Oussalah, Mourad},
  journal={Online Social Networks and Media},
  volume={24},
  pages={100153},
  year={2021},
  publisher={Elsevier}
}

@article{khosla2020contrastive_learning,
  title={Supervised contrastive learning},
  author={Khosla, Prannay and Teterwak, Piotr and Wang, Chen and Sarna, Aaron and Tian, Yonglong and Isola, Phillip and Maschinot, Aaron and Liu, Ce and Krishnan, Dilip},
  journal={Advances in neural information processing systems},
  volume={33},
  pages={18661--18673},
  year={2020}
}

@inproceedings{schroff2015facenet,
  title={FaceNet: A unified embedding for face recognition and clustering},
  author={Schroff, Florian and Kalenichenko, Dmitry and Philbin, James},
  booktitle={Proceedings of the IEEE conference on computer vision and pattern recognition},
  pages={815--823},
  year={2015}
}

@inproceedings{he2018siamese,
  title={A twofold siamese network for real-time object tracking},
  author={He, Anfeng and Luo, Chong and Tian, Xinmei and Zeng, Wenjun},
  booktitle={Proceedings of the IEEE conference on computer vision and pattern recognition},
  pages={4834--4843},
  year={2018}
}

@inproceedings{yadav2024pag,
  title={Pag-llm: Paraphrase and aggregate with large language models for minimizing intent classification errors},
  author={Yadav, Vikas and Tang, Zheng and Srinivasan, Vijay},
  booktitle={Proceedings of the 47th international ACM SIGIR conference on research and development in information retrieval},
  pages={2569--2573},
  year={2024}
}

@inproceedings{agarwal2023perturb,
  title={Towards effective paraphrasing for information disguise},
  author={Agarwal, Anmol and Gupta, Shrey and Bonagiri, Vamshi and Gaur, Manas and Reagle, Joseph and Kumaraguru, Ponnurangam},
  booktitle={European Conference on Information Retrieval},
  pages={331--340},
  year={2023},
  organization={Springer}
}

@inproceedings{chen2021snx,
  title={Self-learn to explain siamese networks robustly},
  author={Chen, Chao and Shen, Yifan and Ma, Guixiang and Kong, Xiangnan and Rangarajan, Srinivas and Zhang, Xi and Xie, Sihong},
  booktitle={2021 IEEE International Conference on Data Mining (ICDM)},
  pages={1018--1023},
  year={2021},
  organization={IEEE}
}

@inproceedings{chen2020simclr,
  title={A simple framework for contrastive learning of visual representations},
  author={Chen, Ting and Kornblith, Simon and Norouzi, Mohammad and Hinton, Geoffrey},
  booktitle={International conference on machine learning},
  pages={1597--1607},
  year={2020},
  organization={PmLR}
}

@inproceedings{maynez2020faithfulness,
  title={On faithfulness and factuality in abstractive summarization},
  author={Maynez, Joshua and Narayan, Shashi and Bohnet, Bernd and McDonald, Ryan},
  booktitle={Proceedings of the 58th annual meeting of the association for computational linguistics},
  pages={1906--1919},
  year={2020}
}

@article{gunel2020supervised,
  title={Supervised contrastive learning for pre-trained language model fine-tuning},
  author={Gunel, Beliz and Du, Jingfei and Conneau, Alexis and Stoyanov, Ves},
  journal={arXiv preprint arXiv:2011.01403},
  year={2020}
}

@inproceedings{chen2024unified,
    title = "Unified Hallucination Detection for Multimodal Large Language Models",
    author = "Chen, Xiang  and
      Wang, Chenxi  and
      Xue, Yida  and
      Zhang, Ningyu  and
      Yang, Xiaoyan  and
      Li, Qiang  and
      Shen, Yue  and
      Liang, Lei  and
      Gu, Jinjie  and
      Chen, Huajun",
    booktitle = "Proceedings of the 62nd Annual Meeting of the Association for Computational Linguistics (Volume 1: Long Papers)",
    year = "2024",
    publisher = "Association for Computational Linguistics",
    doi = "10.18653/v1/2024.acl-long.178",
    pages = "3235--3252",
}

@inproceedings{wei2019eda,
    title = "{EDA}: Easy Data Augmentation Techniques for Boosting Performance on Text Classification Tasks",
    author = "Wei, Jason  and
      Zou, Kai",
    booktitle = "Proceedings of the 2019 Conference on Empirical Methods in Natural Language Processing and the 9th International Joint Conference on Natural Language Processing (EMNLP-IJCNLP)",
    year = "2019",
    publisher = "Association for Computational Linguistics",
    doi = "10.18653/v1/D19-1670",
    pages = "6382--6388",
}

@inproceedings{edunov2018understanding,
    title = "Understanding Back-Translation at Scale",
    author = "Edunov, Sergey  and
      Ott, Myle  and
      Auli, Michael  and
      Grangier, David",
    booktitle = "Proceedings of the 2018 Conference on Empirical Methods in Natural Language Processing",
    year = "2018",
    publisher = "Association for Computational Linguistics",
    doi = "10.18653/v1/D18-1045",
    pages = "489--500",
}

@inproceedings{scialom2021questeval,
  title={QuestEval: Summarization asks for fact-based evaluation},
  author={Scialom, Thomas and Dray, Paul-Alexis and Lamprier, Sylvain and Piwowarski, Benjamin and Staiano, Jacopo and Wang, Alex and Gallinari, Patrick},
  booktitle={Proceedings of the 2021 conference on empirical methods in natural language processing},
  pages={6594--6604},
  year={2021}
}

@inproceedings{honovich2022true,
  title={TRUE: Re-evaluating factual consistency evaluation},
  author={Honovich, Or and Aharoni, Roee and Herzig, Jonathan and Taitelbaum, Hagai and Kukliansy, Doron and Cohen, Vered and Scialom, Thomas and Szpektor, Idan and Hassidim, Avinatan and Matias, Yossi},
  booktitle={Proceedings of the 2022 Conference of the North American Chapter of the Association for Computational Linguistics: Human Language Technologies},
  pages={3905--3920},
  year={2022}
}

@inproceedings{niu2024ragtruth,
  title={Ragtruth: A hallucination corpus for developing trustworthy retrieval-augmented language models},
  author={Niu, Cheng and Wu, Yuanhao and Zhu, Juno and Xu, Siliang and Shum, KaShun and Zhong, Randy and Song, Juntong and Zhang, Tong},
  booktitle={Proceedings of the 62nd Annual Meeting of the Association for Computational Linguistics (Volume 1: Long Papers)},
  pages={10862--10878},
  year={2024}
}

@article{chern2023factool,
  title={FacTool: Factuality Detection in Generative AI--A Tool Augmented Framework for Multi-Task and Multi-Domain Scenarios},
  author={Chern, I and Chern, Steffi and Chen, Shiqi and Yuan, Weizhe and Feng, Kehua and Zhou, Chunting and He, Junxian and Neubig, Graham and Liu, Pengfei and others},
  journal={arXiv preprint arXiv:2307.13528},
  year={2023}
}

@article{wei2024long,
  title={Long-form factuality in large language models},
  author={Wei, Jerry and Yang, Chengrun and Song, Xinying and Lu, Yifeng and Hu, Nathan and Huang, Jie and Tran, Dustin and Peng, Daiyi and Liu, Ruibo and Huang, Da and others},
  journal={Advances in Neural Information Processing Systems},
  volume={37},
  pages={80756--80827},
  year={2024}
}

@inproceedings{min2023factscore,
  title={Factscore: Fine-grained atomic evaluation of factual precision in long form text generation},
  author={Min, Sewon and Krishna, Kalpesh and Lyu, Xinxi and Lewis, Mike and Yih, Wen-tau and Koh, Pang and Iyyer, Mohit and Zettlemoyer, Luke and Hajishirzi, Hannaneh},
  booktitle={Proceedings of the 2023 Conference on Empirical Methods in Natural Language Processing},
  pages={12076--12100},
  year={2023}
}

@inproceedings{sennrich2016improving,
  title={Improving neural machine translation models with monolingual data},
  author={Sennrich, Rico and Haddow, Barry and Birch, Alexandra},
  booktitle={Proceedings of the 54th annual meeting of the association for computational linguistics (volume 1: long papers)},
  pages={86--96},
  year={2016}
}

@inproceedings{lin2026cpil,
  title={Cross Paraphrastic Invariance Learning for Hallucination Detection},
  author={Lin, Shanshan and Hong, Dongsheng and Ju, Sibo and Chen, Chao and Xie, Sihong and Liao, Xiangwen},
  booktitle={ICASSP 2026 IEEE International Conference on Acoustics, Speech and Signal Processing (ICASSP)},
  pages={16662--16666},
  year={2026},
  organization={IEEE}
}

@inproceedings{lin2026cchd,
  title={Constrained Paraphrase Consistency for LLM Hallucination Detection},
  author={Lin, Shanshan and Hong, Dongsheng and Ju, Sibo and Chen, Chao and Zhang, Xi and Liao, Xiangwen},
  booktitle={ICASSP 2026 IEEE International Conference on Acoustics, Speech and Signal Processing (ICASSP)},
  pages={16652--16656},
  year={2026},
  organization={IEEE}
}

@misc{qwen2.5,
    title = {Qwen2.5: A Party of Foundation Models},
    url = {https://qwenlm.github.io/blog/qwen2.5/},
    author = {Qwen Team},
    month = {September},
    year = {2024}
}

@article{llama3,
  title={The llama 3 herd of models},
  author={Grattafiori, Aaron and Dubey, Abhimanyu and Jauhri, Abhinav and Pandey, Abhinav and Kadian, Abhishek and Al-Dahle, Ahmad and Letman, Aiesha and Mathur, Akhil and Schelten, Alan and Vaughan, Alex and others},
  journal={arXiv preprint arXiv:2407.21783},
  year={2024}
}

@article{he2021debertav3,
  title={Debertav3: Improving deberta using electra-style pre-training with gradient-disentangled embedding sharing},
  author={He, Pengcheng and Gao, Jianfeng and Chen, Weizhu},
  journal={arXiv preprint arXiv:2111.09543},
  year={2021}
}

@article{lin2025variational,
  title={Variational Dehallucination via Retrieval-Augmented Prompt Learning},
  author={Lin, Hsin-Yi and Chang, Yu-Che and Chien, Jen-Tzung},
  journal={IEEE Transactions on Audio, Speech and Language Processing},
  volume={34},
  pages={259--271},
  year={2025},
  publisher={IEEE}
}

@article{li2023multi,
  title={A multi-level supervised contrastive learning framework for low-resource natural language inference},
  author={Li, Shu'ang and Hu, Xuming and Lin, Li and Liu, Aiwei and Wen, Lijie and Yu, Philip S},
  journal={IEEE/ACM Transactions on Audio, Speech, and Language Processing},
  volume={31},
  pages={1771--1783},
  year={2023},
  publisher={IEEE}
}

@article{li2024unsupervised,
  title={Unsupervised Disentanglement Learning Model for Exemplar-Guided Paraphrase Generation},
  author={Li, Linjian and Cai, Yi and Wu, Xin},
  journal={IEEE/ACM Transactions on Audio, Speech, and Language Processing},
  volume={32},
  pages={1885--1900},
  year={2024},
  publisher={IEEE}
}

@article{li2024information,
  title={Information dropping data augmentation for machine translation quality estimation},
  author={Li, Shuo and Bi, Xiaojun and Liu, Tao and Chen, Zheng},
  journal={IEEE/ACM Transactions on Audio, Speech, and Language Processing},
  volume={32},
  pages={2112--2124},
  year={2024},
  publisher={IEEE}
}

@article{mao2022fact,
  title={Fact-driven abstractive summarization by utilizing multi-granular multi-relational knowledge},
  author={Mao, Qianren and Li, Jianxin and Peng, Hao and He, Shizhu and Wang, Lihong and Yu, Philip S and Wang, Zheng},
  journal={IEEE/ACM Transactions on Audio, Speech, and Language Processing},
  volume={30},
  pages={1665--1678},
  year={2022},
  publisher={IEEE}
}

@misc{mistralnemo2024,
  author       = {{Mistral AI Team}},
  title        = {Mistral NeMo},
  year         = {2024},
  month        = jul,
  howpublished = {\url{https://mistral.ai/news/mistral-nemo/}},
  note         = {12B open language model developed with NVIDIA}
}

@article{zheng2026afev,
  title={Fact in Fragments: Deconstructing Complex Claims via {LLM}-based Atomic Fact Extraction and Verification},
  author={Zheng, Liwen and Li, Chaozhuo and Liu, Zheng and Huang, Feiran and Jia, Haoran and Ye, Zaisheng and Zhang, Xi},
  journal={Expert Systems with Applications},
  volume={302},
  pages={130572},
  year={2026},
  doi={10.1016/j.eswa.2025.130572},
}

\vfill

\end{document}